\documentclass[10pt,twocolumn,letterpaper]{article}

\usepackage[pagenumbers]{cvpr} 

\usepackage{multirow}
\usepackage[dvipsnames]{xcolor}

\usepackage{booktabs}
\usepackage{pifont}
\definecolor{darkgreen}{HTML}{119B08}
\newcommand{\green}[1]{\textcolor{darkgreen}{#1}}
\newcommand{\blackcheck}{\ding{52}}
\newcommand{\gcheck}{\green{\ding{52}}}
\newcommand{\xcheck}{\ding{56}}
\newcommand{\gxcheck}{\green{\ding{56}}}

\def\method{CheX-GPT\xspace}
\def\testset{MIMIC-500\xspace}

\definecolor{cvprblue}{rgb}{0.21,0.49,0.74}
\usepackage[pagebackref,breaklinks,colorlinks,citecolor=cvprblue]{hyperref}

\def\paperID{8812} 
\def\confName{CVPR}
\def\confYear{2024}

\title{\method: Harnessing Large Language Models \\ for Enhanced Chest X-ray Report Labeling}

\author{
Jawook Gu\textsuperscript{1}
\and
\hspace{-5mm}
Kihyun You\textsuperscript{1}
\and
\hspace{-5mm}
Han-Cheol Cho\textsuperscript{2}
\and
\hspace{-5mm}
Jiho Kim\textsuperscript{2}
\and
\hspace{-5mm}
Eun Kyoung Hong\textsuperscript{2}
\and
\hspace{-5mm}
Byungseok Roh\textsuperscript{2}
\vspace{0.01cm}
\\
\textsuperscript{1}Soombit AI \quad \textsuperscript{2}Kakao Brain\\
{\tt\small \{jawook.gu, kihyun.you\}@soombit.ai},\ 
{\tt\small \{simon.cho, tyler.md, amy.hong, peter.roh\}@kakaobrain.com}
}

\begin{document}
\maketitle
\renewcommand{\arraystretch}{1.05}
\begin{abstract}
Free-text radiology reports present a rich data source for various medical tasks, but effectively labeling these texts remains challenging. Traditional rule-based labeling methods fall short of capturing the nuances of diverse free-text patterns. Moreover, models using expert-annotated data are limited by data scarcity and pre-defined classes, impacting their performance, flexibility and scalability. To address these issues, our study offers three main contributions: 1) We demonstrate the potential of GPT as an adept labeler using carefully designed prompts. 2) Utilizing only the data labeled by GPT, we trained a BERT-based labeler, \method, which operates faster and more efficiently than its GPT counterpart. 3) To benchmark labeler performance, we introduced a publicly available expert-annotated test set, \testset, comprising 500 cases from the MIMIC validation set. Our findings demonstrate that CheX-GPT not only excels in labeling accuracy over existing models, but also showcases superior efficiency, flexibility, and scalability, supported by our introduction of the \testset dataset for robust benchmarking.
Code and models are available at \url{https://github.com/Soombit-ai/CheXGPT}
\end{abstract}
    
\section{Introduction}
\label{sec:intro}

The medical field is rich with chest X-ray (CXR) images, each paired with a free-text radiology report. Accurately labeling these reports offers significant advantages. For instance, labels from reports can streamline the classification of corresponding images, bypassing the need for labor-intensive and costly manual annotation, and setting the stage for automated diagnostic tools \cite{cohen2021gifsplanation}. 

Furthermore, these labels provide an opportunity to employ the Clinical Efficacy (CE) metric \cite{chen2020generating} to gauge the accuracy of reports generated in comparison to the originals. Although many past studies on medical report generation \cite{chen2020generating, wang2023metransformer, nicolson2023improving} have relied on natural language generation (NLG) metrics like BLEU \cite{papineni-etal-2002-bleu} or CIDEr \cite{vedantam2015cider}, these aren't tailored for medical report evaluations \cite{jin2023promptmrg}. A clear example is contrasting the statements ``There is no pneumothorax'' and ``There is pneumothorax.'' While they might score similarly in NLG metrics, their medical meanings are diametrically opposed. Hence, the CE metric, calculated by an automated labeler, presents a more relevant method to assess the quality of medical reports.

\begin{table}[t]
    \centering
    \footnotesize
    \setlength{\tabcolsep}{3pt}
    \begin{tabular}{@{}l|ccc@{}}
    \toprule[1.1pt]
    \multicolumn{1}{c|}{Radiology report} & \scriptsize{Rule-based} & \scriptsize{Fine-tuned} & \scriptsize{Proposed} \\
    \midrule
    (a) \textit{Bibasilar \small{\textbf{atlectasis}} is seen.} & \small{\xcheck} & \small{\gcheck} & \small{\gcheck} \\
    \midrule    
    (b) \textit{..., underlying \small{\textbf{consolidation}} is}  & \multirow{2}{*}{\small{\gcheck}} & \multirow{2}{*}{\small{\xcheck}} & \multirow{2}{*}{\small{\gcheck}} \\
    \textit{difficult to exclude.} & & & \\
    \midrule    
    (c) \textit{A calcified \small{\textbf{granuloma}} is noted in} & \multirow{2}{*}{\small{\xcheck}} & \multirow{2}{*}{\small{\xcheck}} & \multirow{2}{*}{\small{\gcheck}} \\
    \textit{the left mid lung.} & & & \\
    \midrule
    (d) \textit{There is evident \small{\textbf{free gas under}}} & \multirow{2}{*}{\small{\xcheck}} & \multirow{2}{*}{\small{\xcheck}} & \multirow{2}{*}{\small{\gcheck}} \\
    \textit{\small{\textbf{the diaphragm}}, which is a critical ...} & & & \\
    \bottomrule[1.1pt]
    \end{tabular}
    \caption{Four examples of report and corresponding model output. (a) Rule-based labeler, CheXpert \cite{irvin2019chexpert}, fails to detect typo. (b) Fine-tuned labeler, CheXbert \cite{smit2020combining}, suffers generalization failure because of limited training data. Proposed \method offers (c) the flexibility to easily adjust the inclusion of specific words within a pre-defined category, as well as (d) scalability for adding new categories without requiring manual annotation.}
    \vspace{-0.3cm}
    \label{tab:labeler_examples}
\end{table}

Current labelers come with their own set of challenges. Specifically, rule-based models  \cite{irvin2019chexpert, peng2018negbio} often miss the nuanced and varied language patterns in radiology reports, making them especially prone to errors from typos and ambiguities, as illustrated in Table \ref{tab:labeler_examples} (a). A fine-tuned model \cite{smit2020combining}, built upon the rule-based system and trained with limited expert-annotated data, may not generalize well to diverse cases, as shown in Table \ref{tab:labeler_examples} (b). Moreover, the existing labelers are tailored to specific, predefined radiological findings. This implies that altering the definition of findings or introducing a new category requires the creation of new rules or additional human annotation. While there are hundreds of primary radiological findings, previous labelers have tackled only a handful. This highlights the pressing need for a more flexible and scalable approach in developing automated radiograph labelers.

Meanwhile, there have been notable advancements in Large Language Models (LLMs), especially with their human-comparable performance achieved through training large models on massive datasets. Their proficiency isn't just in language comprehension but extends to medical knowledge mastery, positioning LLMs as promising tools in healthcare \cite{kung2023performance, nori2023capabilities, buckley2023accuracy}. As such, efforts have been channeled to convert free-text radiology reports into more desired forms using LLMs \cite{adams2023leveraging, sun2023evaluating, liu2023deid}, alongside proving reliable medical advice based on patient histories \cite{dash2023evaluation}.

In this study, we highlighted the capability of LLMs, such as GPT-3.5 and GPT-4 \cite{openai2023gpt4}, to effectively label radiological findings in free-text CXR reports. Notably, this was achieved without the need for fine-tuning or an annotated dataset. We optimized the LLM's potential by crafting tailored prompts, rich in examples for in-context learning. Subsequently, we mapped the LLM's outputs into 13 pre-defined radiological finding categories.

However, relying solely on LLMs for CXR report labeling isn't the most time or cost-efficient approach. Thus, we introduced a more compact labeler utilizing BERT \cite{devlin2018bert} architecture, trained on pseudo-labels created by LLMs. The resulting model, \method, not only outpaced the LLMs with in-context learning but also marked a significant dip in inference time. Remarkably, the \method's training didn't lean on any human-labeled data; it drew entirely from LLM-labeled data, and still managed to outclass both rule-based and fine-tuned labelers.

Through our research, we discovered a notable deficiency in the field of CXR report labeling: the lack of a benchmark test dataset. While MIMIC-CXR \cite{johnson2019mimic} released only validation set reports, CheXpert \cite{irvin2019chexpert} only provided the labels. To bridge this void and ensure an even playing field for labeler comparisons, we're introducing \testset, a dataset manually annotated by experts from a subset of the MIMIC-CXR validation set. This set consists of binary labels across our 13 defined categories, which pair perfectly with the MIMIC-CXR reports.
This dataset not only lays a crucial foundation for future CXR report annotation endeavors but also supports a broad spectrum of medical imaging research, including automated report generation and CXR image classification, where automated labeling tools are key to enhancing evaluation and analysis.



This work's contributions are summarized as follows:
\begin{itemize}
    \item We showcased LLMs as a superior labeler for free-text CXR reports, relying solely on in-context learning, outpacing conventional supervised models.
    \item By harnessing LLM-derived pseudo-labels, we sculpted a compact and efficient model, \method, that elevated performance benchmarks.
    \item We have introduced an expert-labeled dataset, \testset, to support future annotation works and various medical imaging research endeavors as well.
\end{itemize}

\section{Related Work}
\label{sec:related}

\subsection{Labeling Data with Large Language Models}

Deep learning models require a large amount of labeled data for training.
This annotation process can be costly and time-consuming, particularly in specialized fields like law and medicine.
Previous studies have attempted to address this issue by harnessing LLMs.

In one of the pioneering studies~\cite{wang2021}, GPT-3~\cite{brown2020} was employed as a cost-effective data labeler, resulting in a remarkable reduction in annotation costs, ranging from 50\% to 96\%, across 9 NLP tasks.
Moreover, the authors have also demonstrated that models trained on these pseudo-labeled data often outperformed raw GPT-3.
This was attributed to the predictions made on unlabeled samples serving as a regularization technique for the induced models, consequently enhancing their overall performance.
The emergence of more powerful LLMs such as GPT-3.5 is accelerating this approach.
Their robust capability to follow instructions has significantly reduced the complexity of prompt engineering to obtain high-quality pseudo labels for NLP tasks~\cite{gilardi2023}.

The application of LLMs as zero-shot data labelers extends to specialized domains, including medical documents~\cite{agrawal2022,adams2023leveraging,nori2023capabilities}, legal texts~\cite{savelka2023}, and material science articles~\cite{gupta2022matscibert}.
Our study focuses on a specific branch of the medical domain, CXR reports, incorporating newly introduced radiological findings.

\subsection{Radiology Report Labelers}

\begin{figure}[t]
    \centering
    \includegraphics[width=\linewidth]{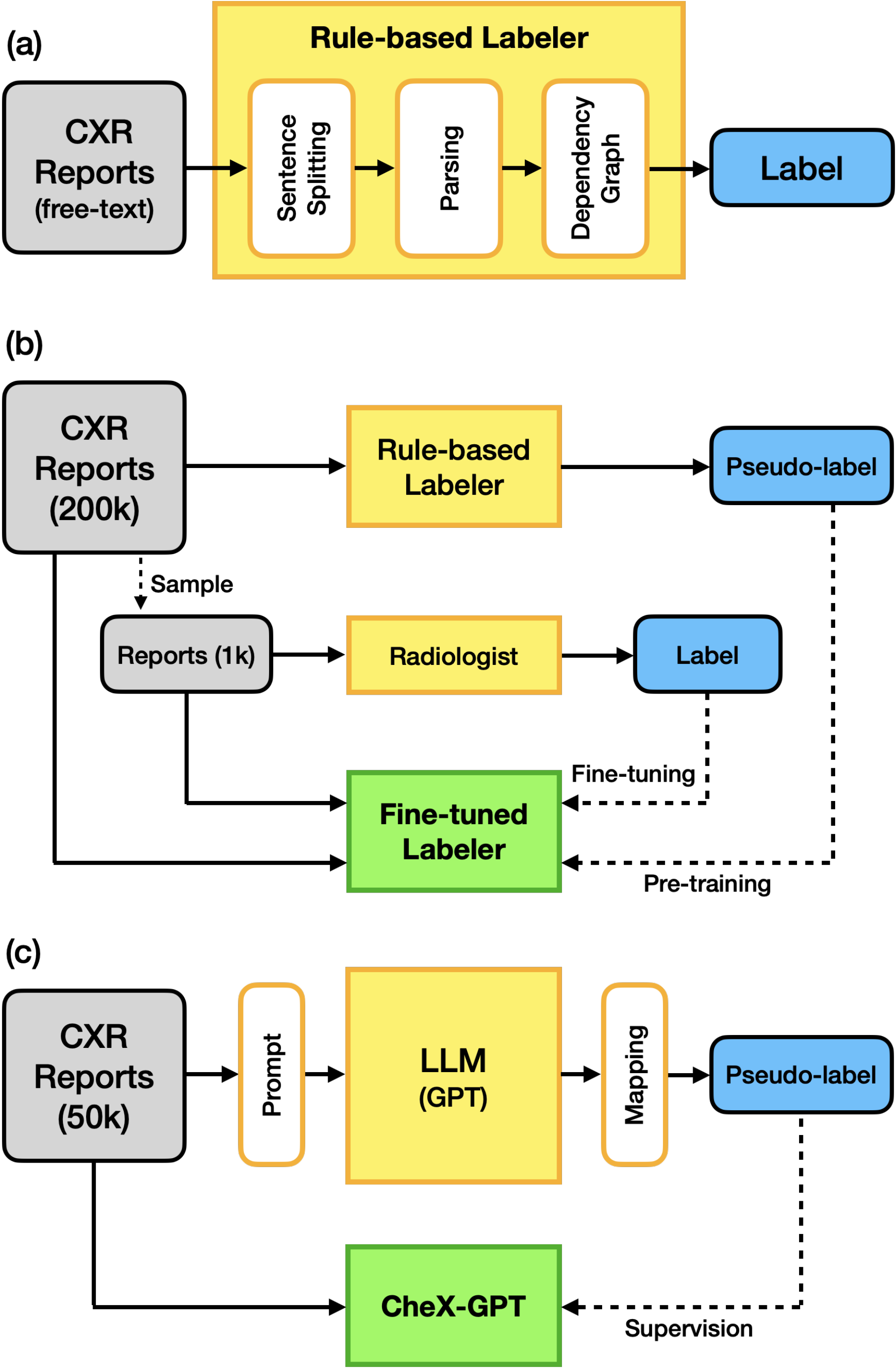}
    \caption{(a) Labeling process for the rule-based labeler. (b) Training framework for the fine-tuned labeler, using a small volume of manually annotated data. (c) Training framework of the proposed \method, employing a large volume of pseudo-labeled data generated by LLM.}
    \vspace{-0.3cm}
    \label{fig:labeler_framework}
\end{figure}

Early studies~\cite{peng2018negbio,irvin2019chexpert} in the field of radiology report classification relied on the expertise of domain experts.
CheXpert~\cite{irvin2019chexpert} enlisted the assistance of professional radiologists to analyze 1,000 radiology reports and formulate a set of rules for creating a rule-based classifier, as illustrated in Figure~\ref{fig:labeler_framework} (a).
CheXbert~\cite{smit2020combining} employs a semi-supervised learning approach, initially pre-training a BERT-based model on the CheXpert dataset, then fine-tuning it on a smaller, manually labeled dataset, as shown in Figure~\ref{fig:labeler_framework} (b). This approach allows CheXbert to surpass the performance of previous rule-based labelers. However, these tools are tailored primarily for the \textit{Impression} section of the CheXpert dataset and confined to specific predefined categories. This specialization limits their effectiveness on different datasets. Additionally, modifying or improving these labelers is challenging: enhancing CheXpert would require a new text corpus, and CheXbert needs extensive new expert annotations.

In contrast, our method, illustrated in Figure~\ref{fig:labeler_framework} (c), leverages a large amount of pseudo-labeled data produced by LLMs instead of a smaller manually labeled dataset. This approach is more adaptable to various datasets. Furthermore, altering the target category is straightforward, requiring only adjustments to the mapping section to align with the user's objectives.

\subsection{Public CXR Datasets}
There has been sustained interest in constructing datasets for CXR research.
Datasets like ChestX-ray8~\cite{wang2017chestx}, CheXpert~\cite{irvin2019chexpert}, MIMIC-CXR~\cite{johnson2019mimic}, PadChest~\cite{bustos2020padchest}, and VinDr-CXR~\cite{nguyen2022vindr} are frequently employed due to their dataset size relative to other datasets.
While many datasets include radiology reports, CheXpert only provides labels and omits the report texts. In contrast, MIMIC-CXR offers both CXR images and reports but lacks structured information such as abnormal radiological findings. Our study introduces expert manual annotations connected to a subset of the MIMIC-CXR validation data. This addition completes a set of reports and labels corresponding to each CXR study, creating a more holistic and structured dataset for testing and aiding future research using CXR report labelers.

\subsection{Evaluation for Multi-modal Model in CXR}
After the introduction of the large-scale image-report datasets such as MIMIC-CXR~\cite{johnson2019mimic} and Open-I~\cite{demner2016preparing}, there has been notable progress in developing multi-modal models integrating both image and text modalities \cite{xue2019improved, li2023dynamic, wang2018tienet}.
Despite these advancements, evaluating such models using NLG metrics poses challenges due to variations in expressions while conveying identical clinical meanings \cite{liao2023deep}.

One potential solution involves employing the CE metric, designed to quantify clinical similarities between ground truth reports and model predictions \cite{miura2020improving, liu2019clinically}. 
While tools like CheXpert~\cite{peng2018negbio} and CheXbert~\cite{irvin2019chexpert} were tailored for the CE metric, the text corpus of CheXpert~\cite{peng2018negbio} built on CheXpert dataset remains unavailable, creating a domain gap with other datasets like MIMIC-CXR and subsequently limiting labeling performance.
In our study, we present a test dataset annotated by medical experts, along with corresponding model-based labelers trained on a sufficient quantity of high-quality pseudo-labels, thereby providing a robust evaluation framework for multi-modal models trained on MIMIC-CXR.
\section{Methods}
\label{sec:methods}

\begin{figure*}[t]
    \centering
    \includegraphics[width=\textwidth]{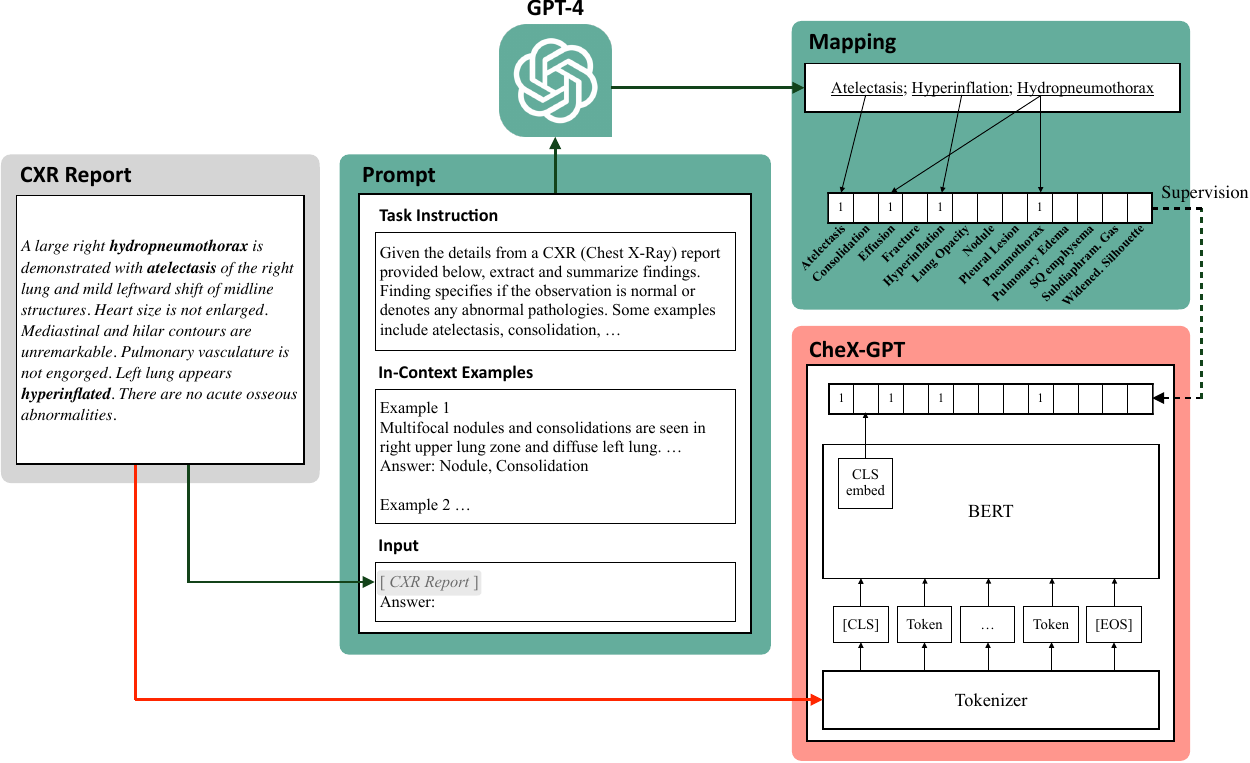}
    \caption{The overall framework mainly consists of the GPT labeler and \method. Initially, a CXR report is placed at the end of the prompt for CXR labeling. GPT-4 then extracts positive findings from the report. The output from GPT-4 is categorized into 13 distinct categories by mapping, and then providing supervision for the BERT-based \method model.}
    \label{fig:chexgpt_model_architecture}
    \end{figure*}

\subsection{Selection of Abnormalities in CXR Reports}

For our study, we focused on 13 key abnormalities frequently identified in CXR reports: atelectasis, consolidation, effusion, fracture, hyperinflation, lung opacity, nodule, pleural lesion, pneumothorax, pulmonary edema, subcutaneous emphysema, subdiaphragmatic gas and widened mediastinal silhouette that includes cardiomegaly \cite{hansell2008fleischner}.
These abnormalities were selected to cover major thoracic illnesses in radiologic reports. 
The accurate identification of the abnormalities is paramount as they can indicate serious underlying conditions, affect treatment decisions, and guide patient management. In the context of automating CXR report annotation, focusing on these abnormalities ensures that the developed model addresses some of the most crucial and frequently encountered findings in the field of radiology.

\subsection{GPT Labeler}

In this study, we engaged GPT-4 \cite{openai2023gpt4}, a state-of-the-art LLM model having medical knowledge, to label potential abnormalities within CXR reports. 

\noindent
\textbf{Prompt} Our methodology's core hinges on a finely crafted prompt structured to guide GPT-4. The prompt is composed of a distinct task instruction, 20 in-context learning examples, and the input CXR report text awaiting labeling, as shown in Figure \ref{fig:chexgpt_model_architecture}. 
The task instruction specifically directs GPT-4 to identify and label all discernable abnormalities present in the provided CXR report. 
The in-context learning examples are carefully curated and refined through an iterative process, where we test GPT-4’s responses, observe patterns of incorrect answers, and subsequently address these errors by incorporating additional in-context examples into the prompt. This methodical approach ensures that each example not only introduces the model to typical terminologies and phrasings found in CXR reports but also actively corrects and fine-tunes its understanding and response accuracy.
Relying on the instruction and examples, the model is tasked with labeling any abnormalities.

\noindent
\textbf{Mapping} Following GPT-4's labeling, a rule-based method is then employed to map the model's outputs to our predefined list of 13 critical abnormalities. By coupling GPT-4's vast language capabilities with our refined prompts and subsequent rule-based mapping, we aim to devise a dependable and efficient method to label CXR reports, spotlighting the 13 key abnormalities of clinical interest.

\subsection{\method}

We utilized the CheXbert architecture~\cite{smit2020combining} for multi-label classification tasks.
The original implementation of CheXbert employs BERT~\cite{devlin2018bert} as its backbone and appends a classification head for each radiological finding, categorizing them into four distinct labels: positive, negative, uncertain, and blank.
In our model, we modified the classification heads for 13 designated categories, simplifying the labels to either `positive' or `not positive', as illustrated in Figure~\ref{fig:chexgpt_model_architecture}.
We trained the model with binary cross-entropy loss.

We trained a model with CheXbert pre-trained weights for efficiency.
Each input report text undergoes tokenization, limiting the maximum number of tokens in each input sequence to 512.
The model underwent 10K steps with a learning rate of $5e^5$ and a batch size of 32.
The learning rate was halved every 4K steps by the LR scheduler.
AdamW optimizer was employed with its default configuration.
We implemented all experiments on PyTorch with an NVIDIA A100 GPU, and the training time takes about one hour.


\subsection{Data}
\label{sec:Data}

\begin{table}[t]\setlength{\tabcolsep}{3pt}
\small
  \centering
  \begin{tabular}{l|c|cc}
    \toprule
    \multirow{2}{*}{Category} &
    {Training set (50k)} & \multicolumn{2}{c}{\testset} \\
    & Longer & Findings & Impression \\
    \midrule
    Atelectasis & 15,461 (30.9\%) & 182 (36.4\%) & 100 (20.0\%) \\
    Consolidation & 5,195 (10.4\%) & 60 (12.0\%)& 51 (10.2\%) \\
    Effusion & 16,583 (33.2\%) & 203 (40.6\%)& 150 (30.0\%) \\
    Fracture & 3,720 (7.4\%) & 49 (9.8\%) & 9 (1.8\%) \\
    Hyperinflation & 3,861 (7.7\%) & 55 (11.0\%) & 17 (3.4\%) \\
    Lung Opacity & 15,206 (30.4\%) & 217 (43.4\%) & 139 (27.8\%) \\
    Nodule & 3,581 (7.2\%) & 58 (11.6\%) & 19 (3.8\%) \\
    Pleural Lesion & 4,130 (8.3\%) & 61 (12.2\%) & 9 (1.8\%) \\
    Pneumothorax & 3,995 (8.0\%) & 26 (5.2\%) & 22 (4.4\%) \\
    Pulmo. Edema & 7,238 (14.5\%) & 147 (29.4\%) & 134 (26.8\%) \\
    SQ Emphysema & 1,813 (3.6\%) & 11 (2.2\%) & 6 (1.2\%)\\
    Subdia. Gas & 868 (1.7\%) & 4 (0.8\%) & 1 (0.2\%) \\
    Widened. Sil. & 12,735 (25.5\%) & 138 (27.6\%) & 48 (9.6\%) \\
    \midrule
    No Abnormality & 14,248 (28.5\%) & 77 (15.4\%) & 165 (33.0\%) \\
    \bottomrule
  \end{tabular}
  \caption{Label distributions of pseudo-labeled training set and hand-crafted test set \testset. For the training set, we selected the longer text segment from either the \textit{Findings} or \textit{Impression} sections.}
  \label{tab:data_statistics}
\end{table}

The primary dataset utilized in this study is MIMIC-CXR~\cite{johnson2019mimic}, comprising 377,110 images, each accompanied by a radiology report. These reports are a valuable source for gleaning insights into various chest abnormalities. For consistency, all reports in this study underwent preprocessing to eliminate superfluous spaces and newlines. 

In the model training phase, we engaged a subset of the MIMIC-CXR dataset, encompassing 50,000 unique CXR reports. 
In selecting this subset, we specifically targeted keywords in the reports to ensure that the training set included a substantial number of positive labels for less common categories such as `fracture' or `subdiaphragmatic gas'.
These were labeled using GPT-4 with specially designed prompts as described previously, and the distributions of the labels are displayed in Table~\ref{tab:data_statistics}. For each report, the longer segment was sampled from either the \textit{Findings} or \textit{Impression} section for training.

For evaluation, we manually annotated 500 reports from the MIMIC-CXR validation set, referring to this subset as \testset. This collection served as a benchmark, allowing us to compare our GPT-4-driven annotation process with human expertise. We compiled unique report sets that included both the \textit{Findings} and \textit{Impression} sections to enable a thorough evaluation of both sections.

Upon examining the \testset column in Table~\ref{tab:data_statistics}, there is a marked difference in label distribution between the \textit{Findings} and \textit{Impression} sections. The \textit{Impression} section exhibits a substantially lower count of positive labels in comparison to \textit{Findings}. 
This is largely due to the characteristics of the \textit{Impression} section, which generally summarizes the \textit{Findings} section by selectively noting only the clinically significant observations or diagnoses inferred from observations. In contrast, the \textit{Findings} section often includes a variety of observations with more descriptive terms, even those with lesser significance, whereas the \textit{Impression} section tends to succinctly document only those findings deemed to be more important.
Consequently, the \textit{Findings} section usually contains more comprehensive information than the \textit{Impression} section. This highlights the importance of an automated labeler for both sections, whereas previous labelers primarily focused on the \textit{Impression} section, where expressions are clearer but less informative.

The process of manual annotation involved a careful review by expert radiologists to ensure accuracy and consistency, with each section being annotated independently. 
Radiology reports, being written in natural language, offer multiple ways to describe identical findings. For instance, `Pleural fluid' denotes fluid in the pleural space, equivalent to pleural effusion. We labeled these varied expressions under the 13 selected abnormalities. Furthermore, the likelihood of findings is often expressed in different terms within the reports. Phrases like `suggesting' indicate high probability, whereas `cannot be excluded' point to a lower probability. Due to the subjectivity inherent in these probability expressions, findings described with low probability were also labeled as `positive’ to maintain objectivity.

\section{Experiments}
\label{sec:experiments}

\subsection{Proposed Labelers}

\begin{table*}[t]\setlength{\tabcolsep}{2pt}
\small
  \centering
  \begin{tabular}{r|ccccc|ccccc}
    \toprule
    \multirow{2}{*}{Category} &
    \multicolumn{5}{c|}{Findings} & \multicolumn{5}{c}{Impression}\\
    & CheXpert & CheXbert & GPT-3.5 & GPT-4 & CheX-GPT & CheXpert & CheXbert & GPT-3.5 & GPT-4 & CheX-GPT \\
    \midrule
    Atelectasis & 83.44 & 82.32 & 95.77 & \textbf{96.63} & \underline{96.34} & 79.76 & 79.52 & 96.48 & \underline{98.02} & \textbf{98.52} \\
    Consolidation & 84.68 & 78.43 & 85.50 & \underline{94.02} & \textbf{96.72} & 80.46 & 73.17 & 95.92 & \underline{96.97} & \textbf{99.01} \\
    Effusion & 91.05 & 90.37 & 90.43 & \underline{97.52} & \textbf{97.77} & 92.04 & 91.93 & 93.47 & \textbf{97.70} & \underline{97.69} \\
    Fracture & 80.43 & 73.56 & 93.88 & \textbf{97.03} & \underline{94.95} & 72.73 & 80.00 & \underline{82.35} & 80.00 & \textbf{88.89} \\
    Lung Opacity & 86.19 & 84.09 & 87.16 & \underline{88.47} & \textbf{88.56} & 89.44 & 89.13 & \underline{93.29} & \textbf{94.85} & 93.06 \\
    Nodule & 65.35 & 70.59 & 80.95 & \textbf{82.09} & \textbf{82.09} & 77.78 & 78.05 & \textbf{87.80} & 82.93 &  \underline{85.71} \\
    Pleural Lesion & 40.00 & 39.53 & \underline{95.08} & \textbf{96.06} & 92.91 & 44.44 & 42.11 & \underline{94.12} & \textbf{94.74} & 90.00 \\
    Pneumothorax & 88.46 & 87.50 & 60.24 & \underline{92.59} & \textbf{92.59} & 93.02 & 92.68 & \textbf{100.00} & \underline{97.78} & 95.45 \\
    Pulmonary Edema & 77.55 & 77.05 & 67.24 & \underline{80.16} & \textbf{83.00} & 84.65 & 85.25 & 74.42 & \underline{85.36} & \textbf{85.47} \\
    Widened Mediastinal Silhouette & 67.76 & 75.94 & 73.97 & \underline{78.32} & \textbf{78.98} & 81.82 & 87.04 & 78.57 & \underline{93.20} & \textbf{94.12} \\ \midrule
    Hyperinflation & - & - & 73.56 & \underline{96.23} & \textbf{99.08} & - & - & 64.00 & \textbf{90.32} & \textbf{90.32} \\
    Subqutaneous Emphysema & - & - & \textbf{90.00} &\textbf{90.00} &\textbf{90.00} & - & - & \textbf{90.91} & 76.92 & \textbf{90.91} \\
    Subdiaphragmatic Gas & - & - & 50.00 & \textbf{85.71} & \textbf{85.71} & - & - & \textbf{100.00} & \textbf{100.00} & \textbf{100.00} \\
    \hline \hline
    Micro-average (10 categories) & 79.68 & 79.76 & 84.73 & \underline{89.83} & \textbf{90.10} & 85.52 & 85.49 & 89.84 & \underline{93.84} & \textbf{93.87} \\
    Macro-average (10 categories) & 76.49 & 75.94 & 83.02 & \underline{90.29} & \textbf{90.39} & 79.61 & 79.89 & 89.64 & \underline{92.15} & \textbf{92.79} \\
    \midrule
    Micro-average (13 categories) & - & - & 84.27 & \underline{90.10} & \textbf{90.48} & - & - & 89.37 & \underline{93.62} & \textbf{93.78}  \\
    Macro-average (13 categories) & - & - & 80.29 & \underline{90.37} & \textbf{90.67} & - & - & 88.56 & \underline{91.45} & \textbf{93.01}  \\
    \bottomrule
  \end{tabular}
  \caption{F1 score of labelers on \testset set. The best and second best results are highlighted using \textbf{bold} and \underline{underline} numbers, respectively. CheX-GPT achieved either the best or the second-best F1 score in most of the diagnosis categories.}
  \label{tab:results}
\end{table*}

The performance of various labelers, including GPT-3.5, GPT-4 and \method, has been evaluated in Table \ref{tab:results}. 

GPT-3.5, an antecedent in the GPT series, demonstrated substantial proficiency in labeling abnormalities in CXR reports.
While it showcased substantial language modeling prowess, its annotations were not as precise as its successor and \method. 

GPT-4, with its more advanced architecture and greater training data, significantly outperformed GPT-3.5. Its enhanced natural language understanding allowed for more nuanced and accurate labeling of CXR reports. As shown in Table \ref{tab:results}, GPT-4's performance metrics in almost all categories showed tangible improvements over GPT-3.5, underscoring the importance of continual model evolution.

Our key achievement, \method, synergies GPT-4's advanced labeling capabilities with the streamlined efficiency of the BERT architecture. While it is trained exclusively on data labeled by GPT-4, \method not only assimilates GPT-4's strengths but also significantly reduces model size. This reduction leads to markedly faster inference times and lower operational costs, without compromising performance. In fact, \method's specialized design for CXR report labeling allows it to perform on par with GPT-4, as evidenced in Table \ref{tab:results}. This balance of efficiency and effectiveness establishes \method as a highly efficient yet powerful tool in our study.


\subsection{Comparative study}

In this section, we compare the labeling accuracy of our \method against the established CheXpert and CheXbert models, as outlined in Table~\ref{tab:results}.

While CheXpert categorizes findings into 14 types and our method uses 13, there are 8 categories that are directly comparable. Moreover, we have matched two additional pairs of categories: `Pleural Other' from CheXpert is matched with our `Pleural Lesion' category, and their `Enlarged Cardiomediastinum' along with `Cardiomegaly' is matched with our `Widened Mediastinal Silhouette' category. This matching process gives us a total of 10 categories for a side-by-side comparison among CheXpert, CheXbert, and \method.

In Table~\ref{tab:results}, the proposed method demonstrated significantly higher performance compared to CheXbert and CheXpert. Upon examining reports where notable differences were observed, we analyzed that these performance disparities primarily stem from two factors. The first cause is attributed to performance errors in CheXbert, predominantly occurring in complex sentences. Examples of these instances are provided in the Appendix.

The second reason is the variance in the definition of categories between this study and CheXbert, specifically in terms of the inclusion of certain words. For instance, the `Pleural Lesion' category shows markedly lower performance in comparative models, mainly because terms like `Costophrenic Angle (CPA) blunting' are not categorized as `Pleural Lesion' by these models. 
CPA blunting, marked by a wider angle at the costophrenic recess in contrast to the acute angle in a normal chest X-ray (CXR), is mostly caused by effusion, although it does not exclusively indicate effusion~\cite{lee2023pleural}. 
Therefore, we categorized CPA blunting as a pleural lesion, a broader classification than just pleural effusion, whereas the comparative models classified it as `No Finding'.
Other keywords contributing significantly to the performance gap include `Granuloma' for `Nodule' and `Deformity' for `Fracture'. More examples are listed in the Appendix.

All models under consideration show improved performance in the \textit{Impression} section compared to the \textit{Findings} section.
This improvement is likely due to the \textit{Impression} section's brevity and clarity, making labeling tasks easier.
Notably, The improvement is particularly pronounced in comparative models because these models were originally developed to target the \textit{Impression} section. 
Additionally, the \textit{Impression} section generally exhibits a lower frequency of key terms such as `CPA Blunting' or `Granuloma', which, as explained earlier, is a key factor in the performance gap observed between the comparative models and our proposed method.

Unlike the ground truth (GT) labels in our study, which categorize the status of findings into `positive' and `not positive', CheXpert and CheXbert differentiate statuses into four categories: `positive', `negative', `uncertain', and `not mentioned'. 
In measuring the performance of these models, we combined `positive' and `uncertain' to compare with the `positive' category of our GT labels. This method was implemented based on our labeling criteria for the test set, where expressions such as `cannot be excluded,' categorized as `uncertain' in existing models, were also treated as `positive'.
When excluding the `uncertain' category and comparing only the `positive' category with our GT labels, we observed a general decline in performance metrics (CheXpert from 77.85 to 76.49, CheXbert from 79.93 to 76.49 for Finding Macro-F1 score). More detailed results on this can be found in the Table of the Appendix.

\subsection{Size of Training Set}

\begin{figure}[t]
    \centering
    \includegraphics[width=\linewidth]{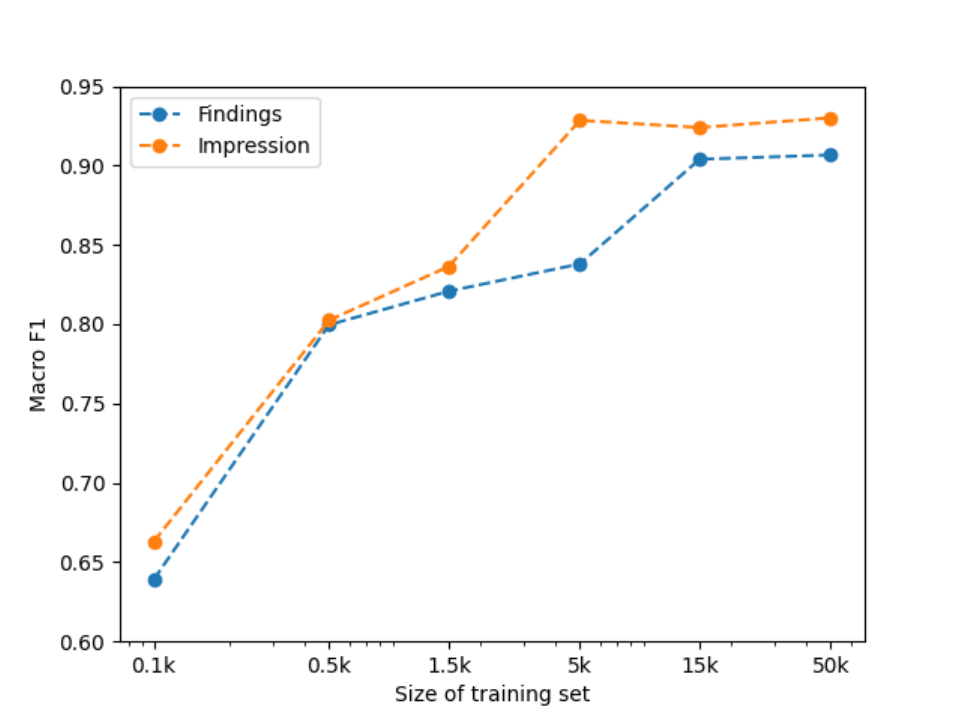}
    \caption{Macro averaged F1 score curve for various training data sizes (log scale). Performance tends to increase as the training data size increases. Performance on \textit{Impressions} is saturated with fewer training samples than on \textit{Findings}.}
    \label{fig:dataset}
\end{figure}

One of the major advantages of the GPT labeler is that, once the prompt is created, the subsequent labeling effort significantly reduces compared to a radiologist annotator, both in terms of time and cost.
In our study, we trained \method models with progressively larger pseudo-labeled datasets to analyze the effect of dataset size on model performance, as illustrated in Figure~\ref{fig:dataset}. Enlarging the training dataset was found to improve performance due to the diversity of expressions included. However, performance reaches a plateau after a certain number of training samples. Notably, this saturation occurs with only 5,000 samples in the \textit{Impression} section, as opposed to 15,000 samples in the \textit{Findings} section, suggesting that \textit{Impression} is a simpler task with limited textual variety. This observation supports the idea that models like CheXbert, which are fine-tuned specifically for the \textit{Impression} section, may be less adaptable when it comes to labeling the \textit{Findings} section.


\subsection{Ablations}

\textbf{In-Context Examples} In-context learning plays a crucial role in enhancing comprehension within specific domains, thereby improving task performance. To demonstrate the effectiveness of in-context examples in our study, we compared performances with and without in-context learning in GPT-3.5 and GPT-4. The results, presented in Table \ref{tab:result_ablation}, are based on the \testset set and are macro-averaged across the 13 categories we defined. The addition of in-context examples generally improves performance, with a notable increase in precision for the \textit{Findings} section. We have documented the cases frequently mistaken when in-context examples were not included in the prompt, detailed in the Appendix.


\begin{table}[t]\setlength{\tabcolsep}{3pt}
\small
  \centering
  \begin{tabular}{l|ccc|ccc}
    \toprule
    \multirow{2}{*}{Model} &
    \multicolumn{3}{c|}{Findings} & \multicolumn{3}{c}{Impression}\\
    & F1 & Precision & Recall & F1 & Precision & Recall \\
    \midrule
    GPT-4 & \textbf{90.37} & \textbf{91.52} & \textbf{90.57} & \textbf{91.45} & \textbf{90.42} & \textbf{93.22} \\
    w.o examples & 83.61 & 82.24 & 90.06 & 89.34 & 87.52 & 93.21 \\
    w.o mapping & 80.33 & 81.39 & 82.21 & 77.79 & 74.37 & 83.88 \\
    \bottomrule
  \end{tabular}
  \caption{Macro-averaged performance metrics for the ablation study of the GPT-4 labeler.}
  \label{tab:result_ablation}
\end{table}

\noindent
\textbf{Mapping} 
In our ablation study, we compared two mapping methods: rule-based mapping and GPT-4 mapping, for classifying findings in CXR reports into 13 predefined categories. As shown in Table \ref{tab:result_ablation}, while GPT-4 mapping has a reasonable degree of accuracy, its performance was still inferior to that of rule-based mapping. The complexity of CXR reports, characterized by indirect term associations and a prevalence of probabilistic language, complicates consistent classification. 
Misclassified examples are available in the Appendix. Based on these observations, we concluded that rule-based mapping was more suitable for our study's objectives.



\begin{table}[t]\setlength{\tabcolsep}{3pt}
\small
  \centering
  \begin{tabular}{l|ccc|ccc}
    \toprule
    \multirow{2}{*}{Model} &
    \multicolumn{3}{c|}{Findings} & \multicolumn{3}{c}{Impression}\\
    & F1 & Precision & Recall & F1 & Precision & Recall \\
    \midrule
    \method & \textbf{90.67} & 91.46 & \textbf{91.20} & \textbf{93.01} & \textbf{93.62} & \textbf{93.27} \\
    w.o pre-train & 89.42 & \textbf{91.53} & 89.62 & 92.42 & 92.75 & 93.07 \\
    \bottomrule
  \end{tabular}
  \caption{Macro-averaged performance metrics for comparing \method with and without the initialization using CheXbert.}
  \label{tab:ablation}
\end{table}

\noindent
\textbf{CheXbert Pretrained Model} The backbone of our \method mirrors has the same architecture of CheXbert~\cite{smit2020combining}. Instead of random initialization, we opt to initialize with the CheXbert model.
This approach, using pre-trained weights, typically yields better performance than models initialized randomly, as shown in Table ~\ref{tab:ablation}. The effectiveness of CheXbert's pre-trained weights likely stems from the similarity of the task.


\section{Discussion}
\label{sec:discussion}

As previously mentioned, our model did not use any data labeled by experts. However, this does not mean that experts were unnecessary in our workflow. In the development process of \method, medical specialists played a crucial role in verifying the accuracy of the LLM's output and identifying errors to update the in-context learning examples during the prompt design phase. Nevertheless, instead of examining all outputs, we found that sampling and checking only a subset of the results was sufficient to enhance the labeler's performance.

To illustrate the efficiency of our proposed methods, we measured the time taken to process a set number of images in the test set by a human expert, compared to the time taken by the GPT-labeler and \method. In practice, it took about 25 hours for a human expert to process 500 images in the test set. If all the 50,000 reports used to train \method were annotated by humans, it would have taken 104 days. In comparison, labeling 50,000 reports took 4 days with GPT-labeler, and 5 minutes with \method, making it much more efficient. Besides the labeling speed, this process also minimizes human error in labeling.

In our comparative study, we reported that a key factor contributing to the performance disparity between existing models and ours is the inclusion of specific keywords in certain categories. In fact, whether such keywords should be included in a particular category can vary depending on the user's intent of the labeler (for example, classifying 'rib deformity' under the fracture category). From this perspective, our model offers a significant advantage not just in terms of performance on the test set, but also in its flexibility to define categories and edit the keywords associated with these categories. By simply modifying the mapping part, users can easily train the model on newly defined categories.

Contrary to CheXpert or CheXbert, which categorize findings into four categories - positive, negative, uncertain, and not mentioned - our model simplifies this to positive or not-positive. This simplification is driven by the essential need to identify positives and the ambiguous distinction between uncertain and positive findings. However, we conducted an additional experiment to test GPT's ability to classify findings into these four categories by modifying the prompt. Since we do not have GT labels for all four categories, we compared the responses of the GPT labeler with those of CheXbert, and the results are in the Table of the Appendix. While this comparison does not provide an exact measure of model performance, it demonstrates GPT's potential to classify findings into diverse statuses.

In terms of scalability, our training framework employing the LLM as a labeler extends its utility beyond merely labeling the presence or absence of findings in CXR reports. It has the capacity to annotate additional details, including the location, certainty, and severity of findings. Furthermore, it can identify comparative expressions within CXR reports, enabling the LLM to determine if a report references previous imaging. While these functionalities have been verified by our team, they are not the focus of this paper and are intended for future investigation.

\section{Conclusion}
\label{sec:conclusion}

The introduction of the \method marks a significant breakthrough in the automated processing of CXR reports. 
By harnessing the advanced capabilities of GPT-4 for initial labeling and further refining this output with a BERT-based framework, our model has outperformed existing systems such as CheXpert and CheXbert. 
\method excels particularly in handling the complex and varied linguistic patterns found in radiology reports, contributing significantly to its heightened accuracy and reliability.

A distinct advantage of our approach is its flexibility in categorizing findings and incorporating specific keywords. This feature greatly enhances the labeling efficiency, allowing \method to be tailored to various clinical needs and making it a versatile tool in medical imaging analysis. Additionally, the model's capability to annotate intricate details in CXR reports, including the location, certainty, and severity of findings, demonstrates its comprehensive analytical scope.

Importantly, the development of \method did not heavily rely on expert-labeled data, significantly reducing the usual resource and time requirements associated with manual annotation. This aspect is especially crucial considering the high costs associated with manual annotation by medical experts. The efficiency and accuracy of \method, therefore, not only make it a practical solution in medical image analysis but also add substantial value in terms of cost-effectiveness.

An integral part of our study was the creation of the \testset test set, a manually annotated dataset that we have made publicly available. This dataset serves as a valuable resource for benchmarking the performance of CXR report labeling systems, including ours. By providing \testset to the research community, we aim to foster further innovation and validation in this field.

In conclusion, \method establishes a new standard in CXR report labeling, blending the power of GPT-4's language processing with the precision of BERT-based architecture. This study not only delivers a powerful tool for medical professionals but also opens new pathways for employing large language models in medical diagnostics and report analysis. The combination of \method's accuracy, efficiency, flexibility, and cost-effectiveness, along with the contribution of the \testset test set, positions it as a transformative development in medical imaging and diagnostics.

{
    \small
    \bibliographystyle{ieeenat_fullname}
    \bibliography{main}

\begin{thebibliography}{38}
\providecommand{\natexlab}[1]{#1}
\providecommand{\url}[1]{\texttt{#1}}
\expandafter\ifx\csname urlstyle\endcsname\relax
  \providecommand{\doi}[1]{doi: #1}\else
  \providecommand{\doi}{doi: \begingroup \urlstyle{rm}\Url}\fi

\bibitem[Adams et~al.(2023)Adams, Truhn, Busch, Kader, Niehues, Makowski, and Bressem]{adams2023leveraging}
Lisa~C Adams, Daniel Truhn, Felix Busch, Avan Kader, Stefan~M Niehues, Marcus~R Makowski, and Keno~K Bressem.
\newblock Leveraging gpt-4 for post hoc transformation of free-text radiology reports into structured reporting: a multilingual feasibility study.
\newblock \emph{Radiology}, 307\penalty0 (4):\penalty0 e230725, 2023.

\bibitem[Agrawal et~al.(2022)Agrawal, Hegselmann, Lang, Kim, and Sontag]{agrawal2022}
Monica Agrawal, Stefan Hegselmann, Hunter Lang, Yoon Kim, and David Sontag.
\newblock Large language models are few-shot clinical information extractors.
\newblock In \emph{Proceedings of the 2022 Conference on Empirical Methods in Natural Language Processing}, pages 1998--2022, 2022.

\bibitem[Brown et~al.(2020)Brown, Mann, Ryder, Subbiah, Kaplan, Dhariwal, Neelakantan, Shyam, Sastry, Askell, et~al.]{brown2020}
Tom Brown, Benjamin Mann, Nick Ryder, Melanie Subbiah, Jared~D Kaplan, Prafulla Dhariwal, Arvind Neelakantan, Pranav Shyam, Girish Sastry, Amanda Askell, et~al.
\newblock Language models are few-shot learners.
\newblock In \emph{Advances in neural information processing systems}, pages 1877--1901, 2020.

\bibitem[Buckley et~al.(2023)Buckley, Diao, Rodman, and Manrai]{buckley2023accuracy}
Thomas Buckley, James~A. Diao, Adam Rodman, and Arjun~K. Manrai.
\newblock Accuracy of a vision-language model on challenging medical cases, 2023.

\bibitem[Bustos et~al.(2020)Bustos, Pertusa, Salinas, and De~La Iglesia-Vaya]{bustos2020padchest}
Aurelia Bustos, Antonio Pertusa, Jose-Maria Salinas, and Maria De~La Iglesia-Vaya.
\newblock Padchest: A large chest x-ray image dataset with multi-label annotated reports.
\newblock \emph{Medical image analysis}, 66:\penalty0 101797, 2020.

\bibitem[Chen et~al.(2020)Chen, Song, Chang, and Wan]{chen2020generating}
Zhihong Chen, Yan Song, Tsung-Hui Chang, and Xiang Wan.
\newblock Generating radiology reports via memory-driven transformer.
\newblock In \emph{Proceedings of the 2020 Conference on Empirical Methods in Natural Language Processing (EMNLP)}, pages 1439--1449, 2020.

\bibitem[Cohen et~al.(2021)Cohen, Brooks, En, Zucker, Pareek, Lungren, and Chaudhari]{cohen2021gifsplanation}
Joseph~Paul Cohen, Rupert Brooks, Sovann En, Evan Zucker, Anuj Pareek, Matthew~P Lungren, and Akshay Chaudhari.
\newblock Gifsplanation via latent shift: a simple autoencoder approach to counterfactual generation for chest x-rays.
\newblock In \emph{Medical Imaging with Deep Learning}, pages 74--104. PMLR, 2021.

\bibitem[Dash et~al.(2023)Dash, Thapa, Banda, Swaminathan, Cheatham, Kashyap, Kotecha, Chen, Gombar, Downing, et~al.]{dash2023evaluation}
Debadutta Dash, Rahul Thapa, Juan~M Banda, Akshay Swaminathan, Morgan Cheatham, Mehr Kashyap, Nikesh Kotecha, Jonathan~H Chen, Saurabh Gombar, Lance Downing, et~al.
\newblock Evaluation of gpt-3.5 and gpt-4 for supporting real-world information needs in healthcare delivery.
\newblock \emph{arXiv preprint arXiv:2304.13714}, 2023.

\bibitem[Demner-Fushman et~al.(2016)Demner-Fushman, Kohli, Rosenman, Shooshan, Rodriguez, Antani, Thoma, and McDonald]{demner2016preparing}
Dina Demner-Fushman, Marc~D Kohli, Marc~B Rosenman, Sonya~E Shooshan, Laritza Rodriguez, Sameer Antani, George~R Thoma, and Clement~J McDonald.
\newblock Preparing a collection of radiology examinations for distribution and retrieval.
\newblock \emph{Journal of the American Medical Informatics Association}, 23\penalty0 (2):\penalty0 304--310, 2016.

\bibitem[Devlin et~al.(2018)Devlin, Chang, Lee, and Toutanova]{devlin2018bert}
Jacob Devlin, Ming-Wei Chang, Kenton Lee, and Kristina Toutanova.
\newblock Bert: Pre-training of deep bidirectional transformers for language understanding.
\newblock \emph{arXiv preprint arXiv:1810.04805}, 2018.

\bibitem[Gilardi et~al.(2023)Gilardi, Alizadeh, and Kubli]{gilardi2023}
Fabrizio Gilardi, Meysam Alizadeh, and Maël Kubli.
\newblock Chatgpt outperforms crowd workers for text-annotation tasks.
\newblock \emph{Proceedings of the National Academy of Sciences}, 120\penalty0 (30):\penalty0 e2305016120, 2023.

\bibitem[Gupta et~al.(2022)Gupta, Zaki, Krishnan, and Mausam]{gupta2022matscibert}
Tanishq Gupta, Mohd Zaki, NM~Anoop Krishnan, and Mausam.
\newblock Matscibert: A materials domain language model for text mining and information extraction.
\newblock \emph{npj Computational Materials}, 8\penalty0 (1):\penalty0 102, 2022.

\bibitem[Hansell et~al.(2008)Hansell, Bankier, MacMahon, McLoud, Muller, and Remy]{hansell2008fleischner}
David~M Hansell, Alexander~A Bankier, Heber MacMahon, Theresa~C McLoud, Nestor~L Muller, and Jacques Remy.
\newblock Fleischner society: glossary of terms for thoracic imaging.
\newblock \emph{Radiology}, 246\penalty0 (3):\penalty0 697--722, 2008.

\bibitem[Irvin et~al.(2019)Irvin, Rajpurkar, Ko, Yu, Ciurea-Ilcus, Chute, Marklund, Haghgoo, Ball, Shpanskaya, et~al.]{irvin2019chexpert}
Jeremy Irvin, Pranav Rajpurkar, Michael Ko, Yifan Yu, Silviana Ciurea-Ilcus, Chris Chute, Henrik Marklund, Behzad Haghgoo, Robyn Ball, Katie Shpanskaya, et~al.
\newblock Chexpert: A large chest radiograph dataset with uncertainty labels and expert comparison.
\newblock In \emph{Proceedings of the AAAI conference on artificial intelligence}, pages 590--597, 2019.

\bibitem[Jin et~al.(2023)Jin, Che, Lin, and Chen]{jin2023promptmrg}
Haibo Jin, Haoxuan Che, Yi Lin, and Hao Chen.
\newblock Promptmrg: Diagnosis-driven prompts for medical report generation.
\newblock \emph{arXiv preprint arXiv:2308.12604}, 2023.

\bibitem[Johnson et~al.(2019)Johnson, Pollard, Greenbaum, Lungren, Deng, Peng, Lu, Mark, Berkowitz, and Horng]{johnson2019mimic}
Alistair~EW Johnson, Tom~J Pollard, Nathaniel~R Greenbaum, Matthew~P Lungren, Chih-ying Deng, Yifan Peng, Zhiyong Lu, Roger~G Mark, Seth~J Berkowitz, and Steven Horng.
\newblock Mimic-cxr-jpg, a large publicly available database of labeled chest radiographs.
\newblock \emph{arXiv preprint arXiv:1901.07042}, 2019.

\bibitem[Kung et~al.(2023)Kung, Cheatham, Medenilla, Sillos, De~Leon, Elepa{\~n}o, Madriaga, Aggabao, Diaz-Candido, Maningo, et~al.]{kung2023performance}
Tiffany~H Kung, Morgan Cheatham, Arielle Medenilla, Czarina Sillos, Lorie De~Leon, Camille Elepa{\~n}o, Maria Madriaga, Rimel Aggabao, Giezel Diaz-Candido, James Maningo, et~al.
\newblock Performance of chatgpt on usmle: Potential for ai-assisted medical education using large language models.
\newblock \emph{PLoS digital health}, 2\penalty0 (2):\penalty0 e0000198, 2023.

\bibitem[Lee and Walker(2023)]{lee2023pleural}
Gregory~M Lee and Christopher~M Walker.
\newblock Pleural thickening: Detection, characterization, and differential diagnosis.
\newblock In \emph{Seminars in Roentgenology}. Elsevier, 2023.

\bibitem[Li et~al.(2023)Li, Lin, Chen, Lin, Liang, and Chang]{li2023dynamic}
Mingjie Li, Bingqian Lin, Zicong Chen, Haokun Lin, Xiaodan Liang, and Xiaojun Chang.
\newblock Dynamic graph enhanced contrastive learning for chest x-ray report generation.
\newblock In \emph{Proceedings of the IEEE/CVF Conference on Computer Vision and Pattern Recognition}, pages 3334--3343, 2023.

\bibitem[Liao et~al.(2023)Liao, Liu, and Spasi{\'c}]{liao2023deep}
Yuxiang Liao, Hantao Liu, and Irena Spasi{\'c}.
\newblock Deep learning approaches to automatic radiology report generation: A systematic review.
\newblock \emph{Informatics in Medicine Unlocked}, page 101273, 2023.

\bibitem[Liu et~al.(2019)Liu, Hsu, McDermott, Boag, Weng, Szolovits, and Ghassemi]{liu2019clinically}
Guanxiong Liu, Tzu-Ming~Harry Hsu, Matthew McDermott, Willie Boag, Wei-Hung Weng, Peter Szolovits, and Marzyeh Ghassemi.
\newblock Clinically accurate chest x-ray report generation.
\newblock In \emph{Machine Learning for Healthcare Conference}, pages 249--269. PMLR, 2019.

\bibitem[Liu et~al.(2023)Liu, Yu, Zhang, Wu, Cao, Dai, Zhao, Liu, Shen, Li, et~al.]{liu2023deid}
Zhengliang Liu, Xiaowei Yu, Lu Zhang, Zihao Wu, Chao Cao, Haixing Dai, Lin Zhao, Wei Liu, Dinggang Shen, Quanzheng Li, et~al.
\newblock Deid-gpt: Zero-shot medical text de-identification by gpt-4.
\newblock \emph{arXiv preprint arXiv:2303.11032}, 2023.

\bibitem[Miura et~al.(2020)Miura, Zhang, Tsai, Langlotz, and Jurafsky]{miura2020improving}
Yasuhide Miura, Yuhao Zhang, Emily~Bao Tsai, Curtis~P Langlotz, and Dan Jurafsky.
\newblock Improving factual completeness and consistency of image-to-text radiology report generation.
\newblock \emph{arXiv preprint arXiv:2010.10042}, 2020.

\bibitem[Nguyen et~al.(2022)Nguyen, Lam, Le, Pham, Tran, Nguyen, Le, Pham, Tong, Dinh, et~al.]{nguyen2022vindr}
Ha~Q Nguyen, Khanh Lam, Linh~T Le, Hieu~H Pham, Dat~Q Tran, Dung~B Nguyen, Dung~D Le, Chi~M Pham, Hang~TT Tong, Diep~H Dinh, et~al.
\newblock Vindr-cxr: An open dataset of chest x-rays with radiologist’s annotations.
\newblock \emph{Scientific Data}, 9\penalty0 (1):\penalty0 429, 2022.

\bibitem[Nicolson et~al.(2023)Nicolson, Dowling, and Koopman]{nicolson2023improving}
Aaron Nicolson, Jason Dowling, and Bevan Koopman.
\newblock Improving chest x-ray report generation by leveraging warm starting.
\newblock \emph{Artificial Intelligence in Medicine}, 144:\penalty0 102633, 2023.

\bibitem[Nori et~al.(2023)Nori, King, McKinney, Carignan, and Horvitz]{nori2023capabilities}
Harsha Nori, Nicholas King, Scott~Mayer McKinney, Dean Carignan, and Eric Horvitz.
\newblock Capabilities of gpt-4 on medical challenge problems.
\newblock \emph{arXiv preprint arXiv:2303.13375}, 2023.

\bibitem[OpenAI(2023)]{openai2023gpt4}
OpenAI.
\newblock Gpt-4 technical report, 2023.

\bibitem[Papineni et~al.(2002)Papineni, Roukos, Ward, and Zhu]{papineni-etal-2002-bleu}
Kishore Papineni, Salim Roukos, Todd Ward, and Wei-Jing Zhu.
\newblock {B}leu: a method for automatic evaluation of machine translation.
\newblock In \emph{Proceedings of the 40th Annual Meeting of the Association for Computational Linguistics}, pages 311--318, Philadelphia, Pennsylvania, USA, 2002. Association for Computational Linguistics.

\bibitem[Peng et~al.(2018)Peng, Wang, Lu, Bagheri, Summers, and Lu]{peng2018negbio}
Yifan Peng, Xiaosong Wang, Le Lu, Mohammadhadi Bagheri, Ronald Summers, and Zhiyong Lu.
\newblock Negbio: a high-performance tool for negation and uncertainty detection in radiology reports.
\newblock \emph{AMIA Summits on Translational Science Proceedings}, 2018:\penalty0 188, 2018.

\bibitem[Savelka(2023)]{savelka2023}
Jaromir Savelka.
\newblock Unlocking practical applications in legal domain: Evaluation of gpt for zero-shot semantic annotation of legal texts.
\newblock In \emph{Proceedings of the Nineteenth International Conference on Artificial Intelligence and Law}, page 447–451, 2023.

\bibitem[Smit et~al.(2020)Smit, Jain, Rajpurkar, Pareek, Ng, and Lungren]{smit2020combining}
Akshay Smit, Saahil Jain, Pranav Rajpurkar, Anuj Pareek, Andrew~Y Ng, and Matthew Lungren.
\newblock Combining automatic labelers and expert annotations for accurate radiology report labeling using bert.
\newblock In \emph{Proceedings of the 2020 Conference on Empirical Methods in Natural Language Processing (EMNLP)}, pages 1500--1519, 2020.

\bibitem[Sun et~al.(2023)Sun, Ong, Kennedy, Tang, Chen, Elias, Lucas, Shih, and Peng]{sun2023evaluating}
Zhaoyi Sun, Hanley Ong, Patrick Kennedy, Liyan Tang, Shirley Chen, Jonathan Elias, Eugene Lucas, George Shih, and Yifan Peng.
\newblock Evaluating gpt-4 on impressions generation in radiology reports.
\newblock \emph{Radiology}, 307\penalty0 (5):\penalty0 e231259, 2023.

\bibitem[Vedantam et~al.(2015)Vedantam, Lawrence~Zitnick, and Parikh]{vedantam2015cider}
Ramakrishna Vedantam, C Lawrence~Zitnick, and Devi Parikh.
\newblock Cider: Consensus-based image description evaluation.
\newblock In \emph{Proceedings of the IEEE conference on computer vision and pattern recognition}, pages 4566--4575, 2015.

\bibitem[Wang et~al.(2021)Wang, Liu, Xu, Zhu, and Zeng]{wang2021}
Shuohang Wang, Yang Liu, Yichong Xu, Chenguang Zhu, and Michael Zeng.
\newblock Want to reduce labeling cost? {GPT}-3 can help.
\newblock In \emph{Proceedings of the 2021 Conference on Empirical Methods in Natural Language Processing (EMNLP)}, 2021.

\bibitem[Wang et~al.(2017)Wang, Peng, Lu, Lu, Bagheri, and Summers]{wang2017chestx}
Xiaosong Wang, Yifan Peng, Le Lu, Zhiyong Lu, Mohammadhadi Bagheri, and Ronald~M Summers.
\newblock Chestx-ray8: Hospital-scale chest x-ray database and benchmarks on weakly-supervised classification and localization of common thorax diseases.
\newblock In \emph{Proceedings of the IEEE conference on computer vision and pattern recognition}, pages 2097--2106, 2017.

\bibitem[Wang et~al.(2018)Wang, Peng, Lu, Lu, and Summers]{wang2018tienet}
Xiaosong Wang, Yifan Peng, Le Lu, Zhiyong Lu, and Ronald~M Summers.
\newblock Tienet: Text-image embedding network for common thorax disease classification and reporting in chest x-rays.
\newblock In \emph{Proceedings of the IEEE conference on computer vision and pattern recognition}, pages 9049--9058, 2018.

\bibitem[Wang et~al.(2023)Wang, Liu, Wang, and Zhou]{wang2023metransformer}
Zhanyu Wang, Lingqiao Liu, Lei Wang, and Luping Zhou.
\newblock Metransformer: Radiology report generation by transformer with multiple learnable expert tokens.
\newblock In \emph{Proceedings of the IEEE/CVF Conference on Computer Vision and Pattern Recognition}, pages 11558--11567, 2023.

\bibitem[Xue and Huang(2019)]{xue2019improved}
Yuan Xue and Xiaolei Huang.
\newblock Improved disease classification in chest x-rays with transferred features from report generation.
\newblock In \emph{Information Processing in Medical Imaging: 26th International Conference, IPMI 2019, Hong Kong, China, June 2--7, 2019, Proceedings 26}, pages 125--138. Springer, 2019.

\end{thebibliography}
}
\clearpage
\setcounter{page}{1}
\setcounter{table}{0}
\renewcommand{\thetable}{\Alph{section}\arabic{table}}

\appendix
\onecolumn
{\centering
        \Large
        \textbf{\thetitle}\\
        \vspace{0.5em}Supplementary Material \\
        \vspace{1.0em}
}

\section{Example-Based Analysis of Comparative Study}
\label{sec:example_analysis}


\subsection{Labeling Discrepancies in Complex Sentences}

The performance errors in CheXbert, which predominantly occur in complex sentences, are a primary reason for the performance gap between CheXbert and the proposed method. The examples in Table~\ref{tab:simple_error_examples} demonstrate instances where CheXbert failed to accurately label findings that the proposed method could label correctly. 
\begin{itemize}
    \item (a), (b): The proposed method correctly identifies `atelectasis' and `effusion' in the reports, whereas CheXbert and CheXpert do not. The complexity of the sentences, such as the presence of negation in example (a) or conjunctions in example (b), likely contributes to the misclassification by the other models.
    \item (c): The proposed method correctly labels `effusion' as negative, while CheXpert and CheXbert erroneously label it as positive. This discrepancy indicates a potential challenge for the previous labelers in distinguishing between resolved conditions and current findings.
    \item (d): The proposed method detects a `fracture' overlooked by CheXbert, potentially because the preceding sentence asserts the lungs are clear, which may have influenced CheXbert`s annotation decision.
    \item (e), (f): CheXpert and CheXbert categorize terms such as `top-normal heart size' or `shift of the mediastinum' as `widened mediastinal silhouette`, unlike the proposed method. While these expressions suggest a variation from the standard normal for heart or mediastinum, they do not necessarily constitute a `widened mediastinal silhouette', as they do not definitively indicate an enlargement of the mediastinum.
\end{itemize}

Besides the examples summarized in Table~\ref{tab:simple_error_examples}, these achievements impacted the quantitative performance detailed in Table~\ref{tab:results}, where our proposed \method managed to attain the highest performance.

\begin{table}[h]
    \centering
    \small
    \setlength{\tabcolsep}{3pt}
    \begin{tabular}{@{}l|c|ccc@{}}
    \toprule[1.1pt]
    \multicolumn{1}{c|}{Radiology report} & Category & CheXpert & CheXbert & \method \\
    \midrule
    (a) \textit{Worsen aeration in the lungs with no effusion and mild bibasilar \textbf{atelectasis}.} & Atelectasis & \small{\xcheck} & \small{\xcheck} & \small{\gcheck} \\
    \midrule
    (b) \textit{..., there has been mild improvement but not complete resolution of the pre-existing} & \multirow{2}{*}{Effusion} & \multirow{2}{*}{\small{\xcheck}} & \multirow{2}{*}{\small{\xcheck}} & \multirow{2}{*}{\small{\gcheck}} \\
    \textit{pulmonary edema, left \textbf{pleural effusion} with atelectasis, and cardiomegaly.} & & & & \\ \midrule
    (c) \textit{Improved right pneumothorax which is now small. \underline{Resolved} right \textbf{pleural effusion}.} & Effusion & \small{\blackcheck} & \small{\blackcheck} & \small{\gxcheck} \\
    \midrule
    (d) \textit{The lungs are clear of consolidation or effusion. All left posterior 7th rib \textbf{fracture}} & \multirow{2}{*}{Fracture} & \multirow{2}{*}{\small{\xcheck}} & \multirow{2}{*}{\small{\xcheck}} & \multirow{2}{*}{\small{\gcheck}} \\
    \textit{ is identified. Atherosclerotic calcifications noted at the aortic arch.} & & & & \\ \midrule
    \multirow{2}{*}{(e) \textit{There is \textbf{top-normal heart size} with tiny left pleural effusion.}} & Widened media- & 
    \multirow{2}{*}{\small{\blackcheck}} & 
    \multirow{2}{*}{\small{\blackcheck}} & 
    \multirow{2}{*}{\small{\gxcheck}} \\
     & stinal silhouette & & & \\
    \midrule
    \multirow{2}{*}{(f) \textit{... and \textbf{marked shift of the mediastinum} and trachea to the left.}} & Widened media- & 
    \multirow{2}{*}{\small{\blackcheck}} & 
    \multirow{2}{*}{\small{\blackcheck}} & 
    \multirow{2}{*}{\small{\gxcheck}} \\
     & stinal silhouette & & & \\
    \bottomrule[1.1pt]
    \end{tabular}
    \caption{Cases comparing the proposed method with CheXpert and CheXbert across various categories. The green color denotes correct labeling, while a \blackcheck mark and a \xcheck mark represent positive and negative annotations by the labeler, respectively. Best view in color.}
    \label{tab:simple_error_examples}
\end{table}

\newpage

\subsection{Categorization Variability Due to Keyword Interpretation}

The variance in category definitions, particularly concerning the inclusion of specific terms, is the second reason for the discrepancy in performance. In Table~\ref{tab:keyword_error_examples}, we see that the proposed method includes certain keywords under broader categories than CheXbert does. 

\begin{itemize}
    \item (a): `Fractured wire' is recognized by CheXpert and CheXbert but not by the proposed method, as our definition of the `fracture' category is exclusively for abnormalities of bony structures.
    \item (b): The proposed method recognizes `fracture` from the term `deformity', understanding the broader implications of structural abnormalities, while the other models do not make this connection.
    \item (c): `Lung opacity' is not identified by the other models when `fibrosis' is mentioned.
    \item (d): The term `granuloma' is not categorized as a `Nodule' by CheXpert and CheXbert. 
    \item (e): The proposed method categorizes `blunting of the right costophrenic angle' under `Pleural lesion', recognizing the broader implications of the term `blunting', which are missed by the other models.
    \item (f): `Pulmonary edema' is recognized by the proposed method when terms like `pulmonary vascular congestion' are present, which could be associated with edema. CheXpert does not appear to make this link, while CheXbert annotates it as `uncertain'.
\end{itemize}

\begin{table}[h]
    \centering
    \small
    \setlength{\tabcolsep}{3pt}
    \begin{tabular}{@{}l|c|ccc@{}}
    \toprule[1.1pt]
    \multicolumn{1}{c|}{Radiology report} & Category & CheXpert & CheXbert & \method \\
    \midrule
    (a) \textit{Three \textbf{fractured} median sternotomy \underline{wires}. The \underline{wire} located third from the top} & \multirow{2}{*}{Fracture} & \multirow{2}{*}{\small{\blackcheck}} & \multirow{2}{*}{\small{\blackcheck}} & \multirow{2}{*}{\small{\gxcheck}} \\
    \textit{has a \textbf{fracture} fragment oriented posteriorly.} & & & & \\
    \midrule
    (b) \textit{Stable \textbf{deformity} along the right lateral rib cage. No acute findings.} & Fracture & \small{\xcheck} & \small{\xcheck} & \small{\gcheck} \\
    \midrule
    (c) \textit{Persistent biapical \textbf{fibrosis} without superimposed acute consolidation.} & Lung opacity & \small{\xcheck} & \small{\xcheck} & \small{\gcheck} \\
    \midrule
    (d) \textit{A calcified \textbf{granuloma} projects over the right lateral mid lung.} & Nodule & \small{\xcheck} & \small{\xcheck} & \small{\gcheck} \\
    \midrule
    (e) \textit{\textbf{Blunting of the right costophrenic angle} may be due to overlying soft tissue, ...} & Pleural lesion & \small{\xcheck} & \small{\xcheck} & \small{\gcheck} \\
    \midrule
    (f) \textit{Compared to prior study, there is increased \textbf{pulmonary vascular congestion}.} & Pulmonary edema & \small{\xcheck} & \small{\gcheck} & \small{\gcheck} \\
    \bottomrule[1.1pt]
    \end{tabular}
    \caption{Discrepancies in Keyword-Based Categorization illustrate the differences in how specific terms are classified by CheXpert, CheXbert, and the proposed method. The green color denotes correct labeling, while a \blackcheck mark and a \xcheck mark represent positive and negative annotations by the labeler, respectively. Best view in color. 
    }
    \label{tab:keyword_error_examples}
\end{table}

\newpage

\section{Case Study on Ablation Study}
\label{sec:example_analysis_ablation}
\setcounter{table}{0}

\subsection{Case Study on the Role of In-Context Learning in LLM Performance}

In-context learning is essential for enhancing the performance of Large Language Models (LLMs) in specialized tasks, as it significantly boosts comprehension. To evaluate the effect of in-context learning examples on GPT-4`s labeling accuracy, we performed a set of experiments with the model operating both with and without these examples. The qualitative outcomes of these tests are presented in Table \ref{tab:result_ablation}. This section delves into a case study analysis based on these results.

Table~\ref{tab:no_example_error_examples} delineates specific cases that underscore the impact of omitting in-context examples, leading to GPT-4`s misclassification. Cases (a) to (d) highlight a pattern where GPT-4`s prompt, lacking in-context examples, tends to misinterpret statements with negations, resulting in inaccurate labeling. The inclusion of in-context examples corrects this issue, showcasing their importance in understanding the context. In contrast, cases (e) to (g) illustrate scenarios where GPT-4 overlooks positive findings, potentially due to the intricate construction of the report`s text. This case study underscores the significance of in-context examples in improving GPT-4`s interpretative accuracy, particularly in handling complex linguistic structures and negations in medical reports.

\begin{table}[h]
    \centering
    \small
    \setlength{\tabcolsep}{3pt}
    \begin{tabular}{@{}l|c|cc@{}}
    \toprule[1.1pt]
    \multicolumn{1}{c|}{Radiology report} & Labels & GPT-4 & w.o examples \\
    \midrule
    (a) \textit{No reaccumulation of pleural fluid or development of \textbf{pneumothorax}.
    } & Pneumothorax & \small{\gxcheck} & \small{\blackcheck} \\
    \midrule
    (b) \textit{Low lung volumes. No definite focal \textbf{consolidation} identified.
    } & Consolidation & \small{\gxcheck} & \small{\blackcheck} \\
    \midrule
    (c) \textit{Severe emphysema without superimposed \textbf{consolidation}.
    } & Consolidation & \small{\gxcheck} & \small{\blackcheck} \\
    \midrule
    \multirow{2}{*}{(d) \textit{No evidence of pneumonia. The \textbf{mediastinum} is not widened.}} & Widened media- & 
    \multirow{2}{*}{\small{\gxcheck}} & 
    \multirow{2}{*}{\small{\blackcheck}} \\
     & stinal silhouette & & \\
    \midrule
    (e) \textit{Improved right \textbf{pneumothorax} which is now small.
    } & Pneumothorax & \small{\gcheck} & \small{\xcheck} \\
    \midrule
    (f) \textit{Diffusely increased interstitial markings compatible with interstitial
    } & \multirow{2}{*}{Pulmonary edema} & \multirow{2}{*}{\small{\gcheck}} & \multirow{2}{*}{\small{\xcheck}} \\
    \textit{\textbf{edema} versus chronic changes.} & & & \\
    \midrule
    (g) \textit{Improvement of multifocal infiltrates but persistent \textbf{densities} in right} & \multirow{2}{*}{Lung opacity} & \multirow{2}{*}{\small{\gcheck}} & \multirow{2}{*}{\small{\xcheck}} \\
    \textit{middle lobe and peripheral lingula.} & & & \\
    \bottomrule[1.1pt]
    \end{tabular}
    \caption{Failure cases of GPT-4 without In-Context Learning. (a)-(d) Negation errors. (e)-(f) Errors on GT-positive findings. The green color denotes correct labeling, while a \blackcheck mark and a \xcheck mark represent positive and negative annotations by the labeler, respectively. Best view in color.}
    \label{tab:no_example_error_examples}
\end{table}

\newpage

\subsection{Case Study on Mapping Methods for CXR Report Classification}

In our ablation study detailed within the paper, we examined two distinct mapping methods for categorizing findings in CXR reports into our 13 pre-established categories: rule-based mapping and GPT-4 mapping. The comparisons, as reflected in Table \ref{tab:result_ablation}, indicate that although GPT-4 mapping achieves a satisfactory level of accuracy, it is still outperformed by rule-based mapping. The intricate nature of CXR reports, often laden with indirect terminology and probabilistic phrases, presents significant hurdles for consistent classification by AI models like GPT-4.

A deeper analysis of GPT-4’s mapping results, which are tabulated in Table~\ref{tab:examples_mapping}, demonstrates that this method not only scored lower in accuracy but also encountered difficulties in deciphering the underlying reasons for certain misclassifications.

\begin{itemize}
    \item (a): GPT-4 mapping erroneously labels `pneumonia' as `Consolidation', whereas rule-based mapping refrains from categorizing it, recognizing the distinction between the terms.
    \item (b): Similarly to case (a), GPT-4 mapping conflates `pneumonia' with `Consolidation'. Additionally, it overlooks `Lung opacity', which is the label applied by rule-based mapping.
    \item (c): Rule-based mapping accurately categorizes a `mass' as `Nodule', a connection missed by GPT-4 mapping, which fails to label it.
    \item (d): GPT-4 mapping incorrectly attributes `Pulmonary edema' to a case mentioning `congestion', neglecting the qualifying phrase `without overt edema' that rule-based mapping correctly interprets as negating the condition.
    \item (e): GPT-4 mapping inappropriately labels the `resolution of pleural effusion' as ongoing `Effusion', a misclassification not present in rule-based mapping, which likely understands `resolution' as indicative of the condition's improvement or absence.
\end{itemize}

The case studies support that rule-based mapping is more aligned with our study's objectives, offering a more definitive and dependable approach to classifying CXR report findings. 
This method allows for greater control over the categorization dictionary and facilitates simpler corrections of misclassifications.

\begin{table}[h]
    \centering
    \small
    \setlength{\tabcolsep}{3pt}
    \begin{tabular}{@{}l|c|c@{}}
    \toprule[1.1pt]
    \multicolumn{1}{c|}{Radiology report} & Rule-based mapping & GPT-4 mapping \\
    \midrule
    (a) \textit{Right middle lobe and lingular \textbf{pneumonia}.
    } & - & Consolidation \\
    \midrule
    (b) \textit{New retrocardiac \textbf{opacity} concerning for \textbf{pneumonia} in the appropriate clinical setting.
    } & Lung opacity & Consolidation \\
    \midrule
    (c) \textit{Left perihilar opacity corresponding to known pulmonary \textbf{mass} again seen.
    } & Nodule & - \\
    \midrule
    (d) \textit{Vascular congestion \underline{without} overt \textbf{edema}.
    } & - & Pulmonary edema \\
    \midrule
    (e) \textit{Interval \underline{resolution} of right \textbf{pleural effusion}.
    } & - & Effusion \\
    \bottomrule[1.1pt]
    \end{tabular}
    \caption{Error Cases of GPT-4 Mapping. GPT-4 mapping (a) mislabels `pneumonia' as `Consolidation', overlooks (b) `Lung opacity' and (c) `Nodule', (d) mislabels `congestion without edema' as `Pulmonary edema', and (e) incorrectly labels resolved effusion as positive.}
    \label{tab:examples_mapping}
\end{table}

\newpage

\section{Experiments on Diverse Status in Labeling}
\label{sec:additional_comparative}
\setcounter{table}{0}

\subsection{Comparative Analysis of Inclusion Versus Exclusion of `Uncertain' Category}

In Section~\ref{sec:Data}, we outlined our GT labeling approach, which categorizes findings into `positive' or `not positive'. This is in contrast to the four-status categorization of CheXpert and CheXbert, which includes `positive', `negative', `uncertain', and `not mentioned'. Our standard practice for comparative analysis involves combining the `positive' and `uncertain' statuses from the comparative labelers, matching them to `positive' in our GT. 
This adjustment is reflective of our labeling criteria for the test set, wherein terms like `cannot be excluded', which are labeled as `uncertain' by CheXpert and CheXbert, are considered `positive' in our dataset.

This section examines the implications of solely aligning the `positive' status from CheXpert and CheXbert with the `positive' category from our GT, providing an alternative perspective on the influence of including or excluding the `uncertain' category.
Table~\ref{tab:results_uncertain} presents the macro-averaged F1 scores for this experiment. Models annotated with an asterisk (*) indicate where the `uncertain' category has been excluded, highlighting the performance impact of this exclusion. 
The comparison revealed that when `uncertain' labels are not counted as `positive', there is an observable decrease in performance metrics. Specifically, for the Finding Macro-F1 score, CheXpert`s performance dropped from 77.85 to 76.49, and CheXbert`s from 79.93 to 76.49.
These results are well aligned with our GT labeling approach.

\begin{table}[h]\setlength{\tabcolsep}{6pt}
\small
  \centering
  \begin{tabular}{r|cc|cc|cc|cc}
    \toprule
    \multirow{2}{*}{Category} &
    \multicolumn{4}{c|}{Findings} & \multicolumn{4}{c}{Impression}\\
    & CheXpert & CheXpert* & CheXbert & CheXbert* & CheXpert & CheXpert* & CheXbert & CheXbert* \\
    \midrule
    Atelectasis & 94.86 & 83.44 & 95.16 & 82.32 & 98.02 & 79.76 & 98.51 & 79.52 \\ 
    Consolidation & 83.33 & 84.68 & 87.59 & 78.43 & 96.15 & 80.46 & 92.31 & 73.17 \\ 
    Effusion & 94.61 & 91.05 & 96.76 & 90.37 & 93.60 & 92.04 & 95.65 & 91.93 \\ 
    Fracture & 82.98 & 80.43 & 77.78 & 73.56 & 72.73 & 72.73 & 80.00 & 80.00 \\ 
    Hyperinflation & 84.10 & 86.19 & 83.52 & 84.09 & 88.81 & 89.44 & 88.81 & 89.13 \\ 
    Lung Opacity & 68.57 & 65.35 & 69.81 & 70.59 & 82.05 & 77.78 & 75.56 & 78.05 \\ 
    Nodule & 44.44 & 40.00 & 46.32 & 39.53 & 47.62 & 44.44 & 45.45 & 42.11 \\ 
    Pleural Lesion & 78.12 & 88.46 & 90.20 & 87.50 & 95.65 & 93.02 & 100.00 & 92.68 \\ 
    Pneumothorax & 89.86 & 77.55 & 91.78 & 77.05 & 90.14 & 84.65 & 91.49 & 85.25 \\ 
    Pulmonary Edema & 57.57 & 67.76 & 60.40 & 75.94 & 77.05 & 81.82 & 80.00 & 87.04 \\ \midrule
    Micro-average & \textbf{80.73} & 79.68 & \textbf{82.22} & 79.76 & \textbf{90.08} & 85.52 & \textbf{90.81} & 85.49 \\ 
    Macro-average & \textbf{77.85} & 76.49 & \textbf{79.93} & 75.94 & \textbf{84.18} & 79.61 & \textbf{84.78} & 79.89 \\
    \bottomrule
  \end{tabular}
  \caption{Macro-averaged F1 Scores on MIMIC-500, demonstrating the impact of treating the `uncertain` category by CheXpert and CheXbert. Models marked with an asterisk (*) consider only `positive` labels as `positive', while models without an asterisk treat both `positive' and `uncertain' as `positive'. Excluding the `uncertain' category results in a divergence from our GT alignment.}
  \label{tab:results_uncertain}
\end{table}

\newpage

\subsection{GPT-4 Labeling Across Extended Status Categories}

This section demonstrates an experiment assessing GPT-4`s ability to classify CXR findings into four distinct statuses: positive, negative, uncertain, and not mentioned. This assessment goes beyond our proposed model`s binary classification of positive versus not-positive findings.
It is crucial to note that the table does not directly compare GPT-4`s performance with ground truth labels for all four statuses, as these labels do not exist. Instead, the comparison attempts to explain GPT-4`s capacity for detailed classification, illustrating its adaptability to other models` standards like CheXbert`s.

For the experiment, we adapted the GPT prompt to match CheXbert`s four-status system and compared GPT-4`s classification performance against CheXbert`s on the MIMIC-500 dataset. We requested GPT-4 to classify the status of CXR findings, incorporating the following instruction into the prompt to guide its responses: ``Status indicates whether the finding is positive, negative, or uncertain. An uncertain status means it is used in cases of speculation such as `cannot be excluded,' `possibly,' `suggest,' or when vague expressions like `unchanged, `stable,' are used.'' Table~\ref{tab:gpt4_mapping_4_status} displays a cross-tabulation of statuses as classified by GPT-4 (rows) and the corresponding classifications by CheXbert (columns) in both the \textit{Findings} and \textit{Impression} sections. Considering the scarce occurrences of specific statuses within particular categories, such as `fracture' labeled as `uncertain', micro-averaged scores are presented to report performance, minimizing the unwanted influence that a few misclassifications could have on macro-averaged scores.


\begin{table}[h!]\setlength{\tabcolsep}{5pt}
\small
  \centering
  \begin{tabular}{r|cccc|cccc}
    \toprule
     &
    \multicolumn{4}{c|}{Findings (CheXbert, \%)} & \multicolumn{4}{c}{Impression (CheXbert, \%)}\\
    GPT-4 & Positive & Negative & Uncertain & Not mentioned & Positive & Negative & Uncertain & Not mentioned \\
    \midrule
    Positive & \textbf{83.37 (802)} & 1.35 (13) & 6.44 (62) & 8.84 (85) & \textbf{81.49 (427)} & 1.34 (7) & 6.68 (35) & 10.50 (55)
    \\ 
    Negative & 2.52 (26) & \textbf{81.36 (838)} & 8.16 (84) & 7.96 (82) & 1.66 (8) & 13.46 (65) & 1.45 (7) & \textbf{83.44 (403)} \\ 
    Uncertain & 27.57 (59) & 10.28 (22) & \textbf{33.18 (71)} & 28.97 (62) & \textbf{39.38 (89)} & 3.10 (7) & 28.32 (64) & 29.20 (66) \\ 
    Not mentioned & 3.69 (103) & 1.00 (28) & 1.79 (50) & \textbf{93.52 (2613)} & 2.04 (77)  & 0.19 (7) & 0.69 (26) & \textbf{97.08 (3657)}
    \\
    \bottomrule
  \end{tabular}
  \caption{Cross-Tabulation of CheXbert and GPT-4 classification concordance. Rows indicate the GPT-4 classifications, while columns show CheXbert's corresponding reclassifications for \textit{Findings} and \textit{Impression} categories. Percentage values denote the frequency of GPT-4's agreement with CheXbert's categories, and numbers in parentheses indicate the total count of instances for each intersecting classification.}
  \label{tab:gpt4_mapping_4_status}
\end{table}


Table~\ref{tab:gpt4_mapping_4_status} demonstrates a strong alignment between GPT-4 and CheXbert in categorizing `positive' and `not mentioned' statuses. 
A disparity arises in the \textit{Impression} section for cases labeled `negative' by GPT-4, which CheXbert often categorizes as `not mentioned'. 
This discrepancy is largely due to GPT-4's interpretation of definitive negations such as `No acute cardiopulmonary process', a phrase frequently written in the \textit{Impression} section. Despite these phrases signifying the absence of acute conditions, GPT-4 labels them as `negative', contrasting with CheXbert's `not mentioned' categorization.

For `uncertain' labels, GPT-4 and CheXbert show considerable misalignment in both \textit{Findings} and \textit{Impression} sections. Upon review, we identified that GPT-4`s default interpretation of `uncertain' diverges from that of CheXbert, particularly for terms such as `unchanged' and `stable`, which GPT-4 classifies as `uncertain', whereas CheXbert often classifies them as `positive'. Conversely, CheXbert alone tends to mark terms indicative of lower severity like `small', `partial', and `mild' as `uncertain'. On top of that, there`s a notable lack of consistency in how terms like `or', `probable', and `reflect' are classified, with both labelers alternating between `positive' and `uncertain'. From our analysis, the only term consistently labeled as `uncertain' across both labelers was `not excluded'.

In conclusion, the comparison with CheXbert has shown that GPT-4`s classification can align with CheXbert for positive, negative, and not mentioned statuses, except negation phrases where GPT-4 tends to label findings as negative. For `uncertain' status,  discrepancies arise from GPT-4`s distinctive interpretation of uncertainty and inherent inconsistencies among labelers. These observations suggest that with tailored prompt tuning and our proposed training approach, a consistent labeler could be established to categorize findings into a variety of statuses according to the user`s requirements. Addressing these issues and refining the labeler`s performance will be part of future work.


\end{document}